\documentclass[acmsmall,screen,nonacm]{acmart}
\usepackage{booktabs}
\usepackage{multirow}
\usepackage{xcolor}
\usepackage{colortbl}
\usepackage{graphicx}
\usepackage{amsmath}
\usepackage{tikz}
\usetikzlibrary{matrix,positioning}
\usetikzlibrary{arrows.meta,calc,decorations.pathreplacing}
\newcommand{\la}[1]{{\textcolor{red}{~[~\textbf{LA}: \textit{#1} ]}}}

\newcommand{\wk}[1]{{\textcolor{red}{~[~\textbf{WK}: \textit{#1} ]}}}

\newcommand{\pooled}{\ensuremath{A_{\mathrm{pool}}}}
\newcommand{\within}{\ensuremath{A_{\mathrm{within}}}}
\newcommand{\cross}{\ensuremath{A_{\mathrm{cross}}}}

\newcommand{\rqbox}[2]{\par\addvspace{3pt}\noindent
{\setlength{\fboxsep}{3pt}%
\fbox{\parbox{\dimexpr\linewidth-2\fboxsep-2\fboxrule\relax}{%
\small\textbf{#1} #2}}}\par\addvspace{3pt}}

\usepackage{pgfplots}
\pgfplotsset{compat=1.17}

\begin{document}

\title[Disentangling Task Difficulty from Run-Level Failure]
{Disentangling Task Difficulty from Run-Level Failure in Agent Failure Prediction}

\author{Mohsen EsfandyariDoulabi}
\email{mesfand@ncsu.edu}
\affiliation{%
  \institution{North Carolina State University}
  \city{Raleigh}
  \state{North Carolina}
  \country{USA}
}

\author{Lawrence Arkoh}
\email{larkoh@ncsu.edu}
\affiliation{%
  \institution{North Carolina State University}
  \city{Raleigh}
  \state{North Carolina}
  \country{USA}
}

\author{Biruk Tadesse}
\email{batadess@ncsu.edu}
\affiliation{%
  \institution{North Carolina State University}
  \city{Raleigh}
  \state{North Carolina}
  \country{USA}
}

\author{Vaishvi Patel}
\email{vpatel34@ncsu.edu}
\affiliation{%
  \institution{North Carolina State University}
  \city{Raleigh}
  \state{North Carolina}
  \country{USA}
}

\author{Mehul Sharma}
\email{msharm34@ncsu.edu}
\affiliation{%
  \institution{North Carolina State University}
  \city{Raleigh}
  \state{North Carolina}
  \country{USA}
}

\author{Marcelo d'Amorim}
\email{mdamori@ncsu.edu}
\affiliation{%
  \institution{North Carolina State University}
  \city{Raleigh}
  \state{North Carolina}
  \country{USA}
}

\author{Wesley Assun\c{c}\~ao}
\email{wguezas@ncsu.edu}
\affiliation{%
  \institution{North Carolina State University}
  \city{Raleigh}
  \state{North Carolina}
  \country{USA}
}

\begin{abstract}
Predicting whether an LLM agent will eventually fail has emerged as a promising direction for supporting intervention during agent's run. Recent approaches report strong predictive performance, often with AUROC values between $0.85$ and $0.94$. However, the predictors are typically trained  combining  agent's runs (also called, trajectories or traces) from many different tasks. 
We hypothesized that the strong performance may come partly from recognizing that some tasks are harder than others, rather than from detecting whether a particular run is heading toward failure. This distinction matters operationally because task-level difficulty supports decisions about where to allocate computation, whereas run-level failure prediction is needed to decide whether to intervene in an ongoing trajectory. To disentangle task-level difficulty from run-level failure signals, we conducted a study to analyze benchmarks containing repeated attempts of the same task by the same LLM agent.
We separate comparisons across different agents and task units from comparisons between successful and failed runs of the same unit. 
Across the evaluated corpora, pooled AUROC is dominated by comparisons across different agent--task units: more than 99.93\% of the underlying positive--negative pairs are cross-unit. Accordingly, predictors that never observe the current run can achieve high pooled performance, including a difficulty oracle with AUROC up to 0.945, while remaining at chance within task. This means that strong pooled performance may be more informative for deciding where to allocate computation than for deciding whether to intervene in a particular run. In contrast, early run-level discrimination is consistently weak across trajectory predictors, published and released monitors, and hidden-state probes, although it improves later in run and is stronger for weaker agents. Consistent with this distinction, under fixed token budgets, task-level allocation outperforms abort-only strategies, while early stopping becomes beneficial only when within-task AUROC reaches roughly 0.84--0.93, far above the 0.50--0.55 range observed for early monitors.
These findings highlight that failure prediction should be evaluated not only by how accurately it predicts outcomes, but by whether the captured signal supports the deployment decision the predictor is intended to make.
\end{abstract}



\maketitle
\flushbottom
\section{Introduction} 

Large Language Model (LLM)-based agents increasingly support multi-step tasks in software engineering and other domains~\cite{liu2024large, jin2024llms}. Unlike conventional LLM applications that primarily generate a response from a fixed input context, LLM-based agents operate through multi-step interactions in which they take actions, invoke tools, and incorporate observations from the environment while attempting to complete a task~\cite{yao2023react,sweagentpaper}. This increased autonomy enables agents to solve complex problems, but it also introduces substantial execution risk. Recent studies report failure rates ranging from $41\%$ to $86.7\%$ across seven agentic frameworks~\cite{mast}. Such failures are costly because an unsuccessful run may consume considerable time, tokens, and computational resources before its eventual outcome becomes known~\cite{du2026survey,bai2026ai,khanna2026zombie}.
These costs have motivated growing interest in \emph{agent failure prediction}: determining, during execution, whether an ongoing agent run is likely to fail~\cite{barke2026agentrx,zhang2025agent, wang2026failfastrestartsmartearlyfailure}. An accurate failure predictor could support early intervention and avoid wasting resources. For example, by terminating an unpromising execution, requesting human assistance, or changing the agent's strategy before additional resources are consumed. 

Existing work commonly formulates the prediction problem as trajectory classification~\cite{paradox,earlyeval,automata}. They collect successful and failed agent trajectories, train predictors on their trajectories, and use the resulting models to estimate the failure risk of new runs while they are still in progress. These predictors are commonly evaluated using the Area Under the Receiver Operating Characteristic Curve (AUROC), which measures how well a predictor discriminates failed executions from successful ones~\cite{paradox,automata}. Within a benchmark and evaluation condition, a common practice is to pool predictions from executions of different task instances and compute AUROC over the resulting successful and failed runs~\cite{automata,paradox,pham2026agentstop}. We refer to this aggregate measure as \emph{pooled AUROC}. High pooled AUROC values are commonly interpreted as evidence that a predictor can identify executions that are likely to fail. However, this interpretation overlooks an important possibility: the predictor may distinguish tasks with different baseline failure rates rather than distinguish successful and failed executions of the \emph{same task}.

This distinction matters because agent outcomes depend on both the task and the particular execution~\cite{bjarnason2026randomness,difficulty,capable}. Some tasks may fail substantially more often than others. When runs from many tasks are pooled together, a predictor can therefore obtain a high AUROC by assigning higher failure scores to tasks with high baseline failure rates and lower scores to tasks with low baseline failure rates~\cite{janes2008accommodating, borkan2019nuanced}. Such a predictor can rank failures above successes across the benchmark even if it cannot distinguish a successful run from a failed run of the \emph{same task}.
Evidence from prior work suggests that this possibility is realistic. A predictor of task success \emph{before execution begins}, using only static task properties, reaches AUROC of $0.841$ on held-out SWE-bench Verified instances~\cite{psychometrics}, within the range reported by systems that observe agent trajectories heavily. This does not imply that trajectories contain no useful information. Rather, it exposes an ambiguity in pooled evaluation: strong discrimination may arise from between-task differences, run-specific execution signals, or both.
Conflating task-level difficulty and run-level failure can therefore misrepresent a predictor's suitability for runtime intervention. 
Task-level difficulty can help determine where to allocate computation, which LLM to use, or whether a task should be attempted. A runtime monitor must answer a different question: \emph{given that the agent is already executing this task, does the evidence from this particular run indicate that it is heading toward failure?} The latter decision requires information that separates different outcomes of the same task.

Our study investigates how strongly the distinction between task-level difficulty and run-level failure affects the evaluation and use of agent failure predictors. Building on the distinction between pooled and conditional discrimination studied in statistics~\cite{pepe2003statistical,janes2009adjusting} and recent evidence of task-level confounds in agent evaluation~\cite{beyondres,bulut2026s}, we ask three questions: (RQ1) how much reported predictive performance reflects differences between tasks rather than the state of the current execution, (RQ2) whether run-specific signals are available early enough to support intervention, and (RQ3) whether separating these signals changes the deployment decisions suggested by failure predictors. To answer these questions, we exploit repeated agent executions, in which a fixed agent configuration attempts the same task multiple times. Because the task and agent remain constant while the execution varies, successful
and failed attempts within the same model--task unit provide a setting for isolating run-level discrimination from between-unit differences. We analyze repeated-attempt corpora from SWE-agent~\cite{sweagentpaper}, SWE-rebench~\cite{swerebench}, LiveClawBench~\cite{liveclawbench}, and Latent Programming Horizons~\cite{silva2026latentprogramminghorizonscoding}, spanning different agent scaffolds, models, and domains. Empirically, we decompose pooled AUROC into within-unit and cross-unit comparisons; compare run-blind baselines with trajectory-based predictors, published and released monitoring systems, and hidden-state probes at online-computable execution prefixes; and measure within-task discrimination while holding the task and acting model fixed. Finally, we evaluate the practical consequences of these signals through replay experiments under fixed token budgets, comparing task-level resource allocation, run-level aborting, and their combination.




Our results reveal a consistent progression across the three research questions. For RQ1, we find that pooled AUROC is dominated by cross-unit comparisons: more than 99.93\% of the underlying positive--negative pairs compare different model--task units, allowing a difficulty oracle to reach pooled AUROC of 0.945 while remaining at 0.500 within task. For RQ2, early run-level discrimination is consistently weak across trajectory-based predictors, published and released monitoring systems, and hidden-state probes, although the signal becomes stronger later in execution and is more pronounced for weaker agents. For RQ3, this distinction changes the practical value of prediction: under fixed token budgets, task-level allocation outperforms abort-only strategies, while run-level monitoring becomes beneficial only when within-task AUROC reaches roughly 0.84--0.93, far above the 0.50--0.55 range observed for early monitors. These results show that high aggregate predictive performance can reflect task-level difficulty without implying effective runtime failure detection, and that separating these signals is necessary to determine which decisions they can support.

Our study makes three main contributions:
\begin{itemize}

\item \textbf{A measurement framework for agent failure prediction.}
We separate task-level difficulty from run-level failure prediction by decomposing pooled AUROC into within-unit and cross-unit components and by evaluating within-task discrimination on repeated executions.

\item \textbf{An empirical characterization of run-level failure signals.}
Across multiple benchmarks, models, agent scaffolds, trajectory predictors, published monitoring systems, and hidden-state probes, we assess when run-specific information becomes available and whether it is strong enough to support intervention.

\item \textbf{Decision-oriented guidance for evaluation and deployment.}
We connect predictive performance to operational use by comparing task-level allocation with run-level aborting under fixed compute budgets, yielding practical guidance for researchers, tool builders, and practitioners on how failure predictors should be evaluated and used.

\end{itemize}

Taken together, our contributions change how failure-prediction performance should be interpreted for LLM agents. A high pooled AUROC can be useful evidence that failures are associated with predictable differences across tasks, but it does not by itself establish that a monitor can recognize which particular execution of a task will fail. The distinction matters operationally: task-level information supports decisions about \emph{where to spend computation}, whereas run-level information supports decisions about \emph{whether to stop the execution currently in progress}. 

\section{Motivating Example}
\label{sec:example}

Consider two SWE-bench Verified tasks from Django, each attempted ten times by the same agent and model. This gives $20$ runs in total. The agent solves
\texttt{django-10914} nine times but solves \texttt{django-10973} only once. Now consider a predictor that never observes the trajectory and simply assigns each task its observed failure rate: $0.1$ for every run of the first task and $0.9$ for every run of the second. Evaluated across all $20$ runs, this predictor achieves AUROC $0.900$. In either task, however, its AUROC is exactly $0.500$ because every attempt at the same task receives the same score. The predictor therefore distinguishes which task is harder, but not which particular run will fail.
At benchmark scale, this distinction becomes substantial: in C4-L (presented in Table~\ref{tab:corpora}), only about $0.07\%$ of the comparisons underlying pooled AUROC come from runs of the same task. This helps explain why strong pooled performance can coexist with weak run-level discrimination, and why task-level information may be more useful for allocating computation than for deciding whether to abort the current run.

\section{Related Work}
\label{sec:related-work}

Research on agent failure prediction spans three questions: what makes a task likely to fail before execution, what an ongoing trajectory reveals about its eventual outcome, and whether acting on a prediction improves system behavior. These questions rely on different information and support different decisions. Several runtime predictors evaluate executions from multiple tasks together~\cite{automata,paradox,wang2026failfastrestartsmartearlyfailure}, potentially mixing task-level difficulty with run-specific failure signals. We review these lines of work to clarify what existing evaluations establish about run-level monitoring.

\subsection{Task difficulty as a source of predictive performance}

Pre-execution prediction is well established in software engineering, from defect-risk estimation using static software features~\cite{esfandyaridoulabi2025adaptive} to LLM reliability analyses showing sensitivity to task difficulty and repeated evaluation~\cite{sun2026evaluating}. More directly, Ge et al.~\cite{psychometrics} estimate agent task difficulty using Item Response Theory (IRT) and task features. Their model reaches AUROC $0.841$ on held-out SWE-bench Verified tasks, while a problem-statement-only variant reaches $0.786$, showing that substantial discrimination is possible before observing the current execution. Krsteski and Meyer~\cite{difficulty} similarly predict IRT-derived difficulty across 17 agentic benchmarks and show that aggregate AUC can remain high even when task-level scores do not distinguish individual outcomes. Our work, in contrast, asks whether high aggregate failure-prediction performance also captures run-specific information within repeated attempts of the same model--task pair, and whether task-level and run-level signals support different deployment decisions.

Mehtiyev and Assun\c{c}\~ao~\cite{beyondres} further show that trajectory-based associations depend on what is held fixed. Across 9,374 coding-agent trajectories, failed runs are longer when outcomes are compared within an agent, whereas the relationship often reverses when different agents are compared on the same task. Their results highlight the importance of conditioning when interpreting trajectory-level associations. We extend this idea by holding both the acting model and task fixed, allowing direct measurement of discrimination between successful and failed executions of the same model--task unit.
Bulut~\cite{bulut2026s} provides the closest measurement precedent to
our study. In LLM reasoning trajectories, Bulut shows that pooled
correctness-prediction AUROC can largely reflect problem difficulty rather
than attempt-specific information, motivating question-only baselines and
within-problem evaluation. Our work studies the analogous issue in runtime
failure prediction for tool-using LLM agents. We evaluate repeated
model--task executions across multiple agent benchmarks, behavioral
predictors, published and released monitoring systems, and hidden-state
probes, and then examine whether task-level and run-level information support
different deployment decisions under fixed compute budgets.

\subsection{Failure prediction during agent execution}

Recent systems predict eventual success or failure from partial or complete agent trajectories. Cho et al.~\cite{automata} abstract trajectories into finite-state representations and report full-trajectory AUROC of $0.941$ on Tau2-Bench telecom and $0.799$ on SWE-agent. Shi et al.~\cite{earlyeval} train predictors from partial trajectories and report 13--26\% fewer execution steps while maintaining high prediction accuracy. Ruan et al.~\cite{doomed} instead use hidden-state probes to trigger early aborts, substantially reducing generated tokens on TextCraft and WebShop. 
Wang et al.~\cite{wang2026failfastrestartsmartearlyfailure} propose FailFast--RestartSmart, which predicts failure from partial trajectories and restarts unpromising runs, reducing token usage by 14.6--20.4\% at a 5\% false-positive rate. Their training objective includes successful--failed pairs from the same task to control instance difficulty. Our study addresses a complementary evaluation question: how much of a monitor's measured discrimination remains when successful and failed executions are compared within the same model--task unit.

\subsection{Evaluation metrics and deployment decisions}

Raghu et al.~\cite{raghu2026proper} show that AUROC does not assess whether predicted success probabilities remain meaningful across trajectory prefixes and propose a trajectory proper score. Their concern is calibration across time; ours is complementary. We instead ask what source of variation produces the observed discrimination by separating comparisons across model--task units from comparisons between successful and failed executions of the same unit. Prediction quality also does not guarantee useful intervention. Vasudev et al.~\cite{paradox} show that a critic with strong offline failure-prediction performance can reduce task success when interventions disrupt trajectories that would otherwise succeed. This motivates evaluating monitors not only by predictive performance, but also by whether the captured signal supports the intended deployment decision.

\subsection{Synthesis and Research Gap}
Taken together, prior work shows that task difficulty is predictable, trajectories can contain failure-related information, and prediction-guided intervention can help or harm depending on how that information is used. What remains unclear is whether strong aggregate failure-prediction performance reflects task/model difficulty or genuine discrimination between successful and failed executions of the same model--task unit. We address this gap by decomposing pooled AUROC into within-unit and cross-unit components, evaluating predictors at online-computable prefixes, and testing whether run-level signals improve decisions under a fixed compute budget.

\section{Study Design}
\label{sec:design}

Following the Goal--Question--Metric (GQM) framework~\cite{basili1994goal}, the \textbf{goal} of our study is to analyze failure prediction for LLM agents \textbf{for the purpose of} understanding what reported predictive performance captures and whether the captured information supports useful deployment decisions, \textbf{with respect to} the distinction between task-level difficulty and run-level failure, \textbf{from the perspective of} researchers, tool builders, and practitioners, \textbf{in the context of} repeated-attempt agent benchmarks. We define a unit as a $(\text{model},\text{task})$ pair. We use \emph{within-task AUROC} as shorthand for discrimination within a fixed model--task unit, so both the task and acting model are held fixed. The following sections present the research questions (RQs) and details of each step of our study. 
Figure~\ref{fig:design} summarizes the study design and the relationship between the three research questions.


\subsection{Research Questions}
The gaps observed in related work motivate three RQs:

\begin{itemize}
    \item[\textbf{RQ1}] (\emph{Estimand}) \textit{What does pooled failure-prediction
    AUROC measure in benchmarks that record repeated attempts?}
    We examine the contributions of within-unit and cross-unit comparisons
    and assess how much pooled discrimination is achievable without
    observing the current execution.

    \item[\textbf{RQ2}] (\emph{Run-level signal}) \textit{How much run-specific
    information is available at online-computable prefixes, across
    predictors, models, corpora, and hidden-state probes?}
    We compare successful and failed attempts with the task and model
    held fixed to establish the strength and timing of run-level
    discrimination.

    \item[\textbf{RQ3}] (\emph{Deployment}) \textit{Does separating task-level
    from run-level discrimination change intervention and
    resource-allocation conclusions?}
    We compare task allocation, trajectory-triggered aborting, and their
    combination under fixed token budgets, and examine when monitoring
    provides additional value.
\end{itemize}

%
\definecolor{oiBlue}{HTML}{0072B2}
\definecolor{oiVerm}{HTML}{D55E00}
\definecolor{oiGreen}{HTML}{009E73}
\definecolor{oiPurple}{HTML}{CC79A7}
\definecolor{oiSky}{HTML}{56B4E9}
\definecolor{oiOrange}{HTML}{E69F00}
\definecolor{oiYellow}{HTML}{F0E442}
\colorlet{sdSucc}{oiBlue}               
\colorlet{sdFail}{oiVerm}               
\colorlet{sdWithin}{oiGreen!80!black}   
\colorlet{sdCross}{oiPurple!85!black}   
\colorlet{sdWithinTx}{oiGreen!65!black} 
\colorlet{sdCrossTx}{oiPurple!60!black}
\colorlet{sdLine}{black!70}
\colorlet{sdNeutral}{black!22}           
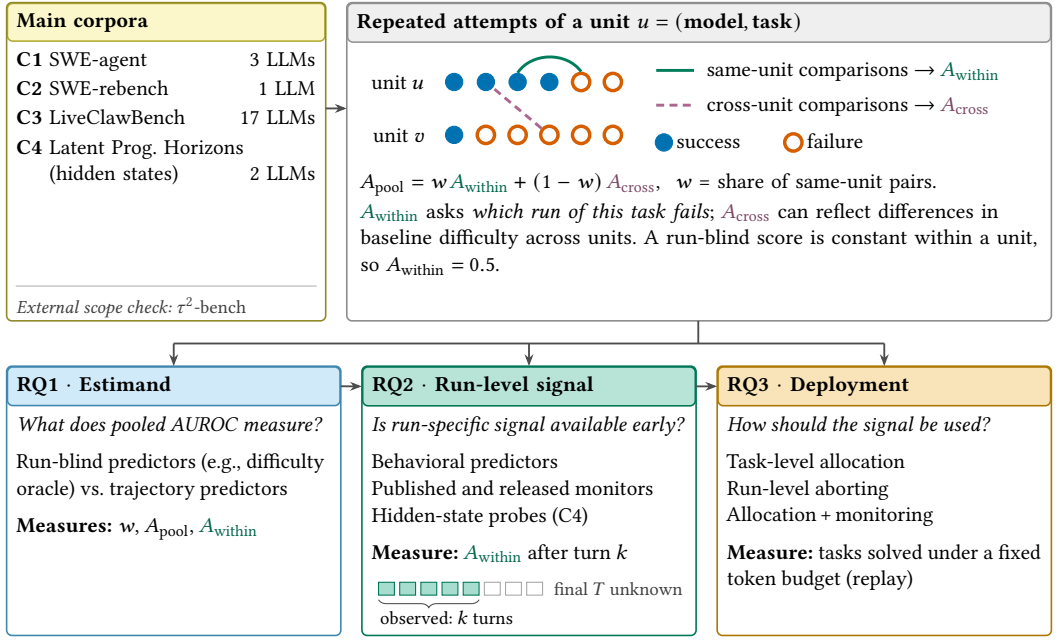
\begin{figure}[t]
\centering
\begin{tikzpicture}[
  font=\footnotesize,
  >={Stealth[length=4.5pt]},
  hd/.style={rounded corners=2pt, line width=0.6pt, align=left, inner sep=3.5pt,
             anchor=north west, font=\footnotesize\bfseries,
             draw=#1!75!black, fill=#1!28},
  body/.style={anchor=north west, inner sep=3.5pt, align=left},
  succ/.style={circle, fill=sdSucc, draw=sdSucc, inner sep=0pt, minimum size=6.5pt},
  fail/.style={circle, fill=white, draw=sdFail, line width=1.3pt, inner sep=0pt, minimum size=6.5pt},
  flow/.style={->, sdLine, line width=0.7pt},
  wline/.style={sdWithin, line width=1.1pt},
  cline/.style={sdCross, line width=1.1pt, dash pattern=on 3pt off 2pt},
]
\def\colw{4.4}   
\def\gap{0.3}
\def\Aw{4.2}     
\def\Bx{4.5}     
\def\Bw{9.3}     
\newcommand{\Wc}{\textcolor{sdWithinTx}{\within}}
\newcommand{\Cc}{\textcolor{sdCrossTx}{\cross}}

\node[hd=oiYellow, text width=\Aw cm-7pt] (hA) at (0,0) {Main corpora};
\node[body, text width=\Aw cm-7pt] (bA) at (hA.south west) {%
  \textbf{C1} SWE-agent\hfill 3 LLMs\\[1pt]
  \textbf{C2} SWE-rebench\hfill 1 LLM\\[1pt]
  \textbf{C3} LiveClawBench\hfill 17 LLMs\\[1pt]
  \textbf{C4} Latent Prog.\ Horizons\\
  \phantom{\textbf{C4}} (hidden states)\hfill 2 LLMs};

\node[hd=sdNeutral, text width=\Bw cm-7pt] (hB) at (\Bx,0)
  {Repeated attempts of a unit $u=(\text{model},\text{task})$};
\def\dx{0.42}
\def\yu{-1.05}
\def\yv{-1.75}
\def\dotx{\Bx+1.0}
\node[anchor=east] at (\dotx+0.15,\yu) {unit $u$};
\node[anchor=east] at (\dotx+0.15,\yv) {unit $v$};
\foreach \i/\s in {1/succ,2/succ,3/succ,4/succ,5/fail,6/fail}
  \node[\s] (u\i) at (\dotx+\i*\dx,\yu) {};
\foreach \i/\s in {1/succ,2/fail,3/fail,4/fail,5/fail,6/fail}
  \node[\s] (v\i) at (\dotx+\i*\dx,\yv) {};
\draw[wline] (u3.north) to[bend left=55] (u5.north);   
\draw[cline] (v4) -- (u2);                             
\def\lx{\dotx+2.95}
\node[anchor=west] at (\lx,\yu+0.14)
  {\tikz[baseline=-0.6ex]\draw[wline](0,0)--(0.5,0);\;
   same-unit comparisons $\rightarrow$ \Wc{}};
\node[anchor=west] at (\lx,\yu-0.32)
  {\tikz[baseline=-0.6ex]\draw[cline](0,0)--(0.5,0);\;
   cross-unit comparisons $\rightarrow$ \Cc{}};
\node[anchor=west] at (\lx,\yv-0.1)
  {\tikz[baseline=-0.6ex]\node[succ]{};\,success\qquad
   \tikz[baseline=-0.6ex]\node[fail]{};\,failure};
\node[body, text width=\Bw cm-0.35cm] (fB) at (\Bx+0.05,\yv-0.38) {%
  $\pooled = w\,\Wc + (1-w)\,\Cc$, \ $w$ = share of same-unit pairs.\\[1pt]
  \Wc{} asks \emph{which run of this task fails}; \Cc{} can reflect differences in baseline difficulty across units. A run-blind score is constant within
  a unit, so $\within=0.5$.};

\path let \p1=(fB.south), \p2=(bA.south) in
  \pgfextra{\pgfmathparse{min(\y1,\y2)}\xdef\tb{\pgfmathresult pt}};
\coordinate (topbot) at (0,\tb);
\coordinate (scope) at ($(topbot)+(0,-0.48)$);   
\coordinate (rq) at ($(scope)+(0,-0.6)$);
\draw[oiYellow!60!black, line width=0.6pt, rounded corners=2pt] (hA.north west) rectangle (hA.east |- scope);
\draw[black!45, line width=0.6pt, rounded corners=2pt] (hB.north west) rectangle (hB.east |- scope);
\draw[flow] (hA.east |- 0,-1.4) -- (hB.west |- 0,-1.4);
\draw[black!30, line width=0.4pt] ($(hA.west |- topbot)+(0.12,-0.02)$) -- ($(hA.east |- topbot)+(-0.12,-0.02)$);
\node[anchor=north west, inner sep=3.5pt, text=black!75, font=\scriptsize] at ($(hA.west |- topbot)+(0,-0.04)$)
  {\textit{External scope check:} $\tau^2$-bench};

\newcommand{\rqbody}[1]{\parbox[t]{\dimexpr\colw cm-7pt\relax}{\raggedright #1}}
\foreach \k/\t/\c [count=\n from 0] in {1/Estimand/oiSky, 2/Run-level signal/oiGreen, 3/Deployment/oiOrange}{%
  \pgfmathsetmacro{\xx}{\n*(\colw+\gap)}
  \node[hd=\c, text width=\colw cm-7pt] (hR\k) at ($(rq)+(\xx,0)$) {RQ\k\ $\cdot$ \t};
}
\node[body] (bR1) at (hR1.south west) {\rqbody{%
  \textit{What does pooled AUROC measure?}\\[4pt]
  Run-blind predictors (e.g., difficulty oracle) vs.\ trajectory predictors\\[4pt]
  \textbf{Measures:} $w$, \pooled{}, \Wc{}}};
\node[body] (bR2) at (hR2.south west) {\rqbody{%
  \textit{Is run-specific signal available early?}\\[4pt]
  Behavioral predictors\\
  Published and released monitors\\
  Hidden-state probes (C4)\\[4pt]
  \textbf{Measure:} \Wc{} after turn $k$}};
\node[body] (bR3) at (hR3.south west) {\rqbody{%
  \textit{How should the signal be used?}\\[4pt]
  Task-level allocation\\
  Run-level aborting\\
  Allocation\,+\,monitoring\\[4pt]
  \textbf{Measure:} tasks solved under a fixed token budget (replay)}};

\begin{scope}[shift={($(bR2.south west)+(0.22,-0.3)$)}]
  \foreach \i in {1,...,5} \draw[oiGreen!75!black, fill=oiGreen!35] (\i*0.28-0.28,0) rectangle +(0.21,0.17);
  \foreach \i in {6,...,8} \draw[black!35, fill=white] (\i*0.28-0.28,0) rectangle +(0.21,0.17);
  \draw[decorate, decoration={brace, amplitude=3pt, mirror}, sdLine] (0,-0.06) -- (1.33,-0.06);
  \node[font=\scriptsize, anchor=north west, inner sep=1pt] (tl) at (-0.02,-0.16) {observed: $k$ turns};
  \node[font=\scriptsize, text=black!75, anchor=west] at (2.2,0.085) {final $T$ unknown};
\end{scope}

\path let \p1=(bR1.south), \p2=(tl.south), \p3=(bR3.south) in
  \pgfextra{\pgfmathparse{min(\y1,\y2,\y3)-3}\xdef\rb{\pgfmathresult pt}};
\foreach \k/\c in {1/oiSky,2/oiGreen,3/oiOrange}
  \draw[\c!75!black, line width=0.6pt, rounded corners=2pt] (hR\k.north west) rectangle (hR\k.east |- 0,\rb);

\coordinate (bus) at ($(rq)+(0,0.3)$);
\draw[sdLine, line width=0.7pt] (\Bx+\Bw/2,0 |- scope) -- (\Bx+\Bw/2,0 |- bus);
\draw[sdLine, line width=0.7pt] (hR1.north |- bus) -- (hR3.north |- bus);
\foreach \k in {1,2,3} \draw[flow] (hR\k.north |- bus) -- (hR\k.north);
\draw[flow] (hR1.east) -- (hR2.west);
\draw[flow] (hR2.east) -- (hR3.west);
\end{tikzpicture}
\caption{\textbf{Overview of the study design.} Repeated attempts within fixed model-task units allow pooled AUROC to be separated into \textbf{same-unit (\within{})} and \textbf{cross-unit (\cross{})} discrimination. We use this distinction to examine what pooled failure prediction measures, when run specific signal emerges, and how the resulting information supports deployment decisions.}
\label{fig:design}
\end{figure}


\subsection{Corpora}
\label{sec:corpora}

The behavioral analyses of our study use three public corpora with repeated attempts and permissive licences. They differ in benchmark, scaffold, model family, domain, and failure rate. Table~\ref{tab:corpora} summarizes their repeat structure. Failure rate is the fraction of trajectories labelled as failures; a mixed unit contains at least one success and one failure; and the number of within-unit positive--negative pairs is $\sum_u |S_u||F_u|$.
Observed mixedness depends on the number of attempts recorded for a unit. For latent success probability $q$ and $r$ attempts, the probability of observing both outcomes is
$1-q^r-(1-q)^r.$
Thus, the mixed-outcome rates in Table~\ref{tab:corpora} describe the observed repeat structure rather than intrinsic task flakiness.


\begin{table*}[t]
\caption{Corpora used in the study. C1--C3 support the main behavioral analyses, while C4-L and C4-Q support the hidden-state analyses. Mixed-outcome is the share of units containing both a successful and a failed run.}
\label{tab:corpora}

\centering
\scriptsize
\setlength{\tabcolsep}{6pt}

\begin{tabular}{lccccc}
\toprule
 & \textbf{C1} & \textbf{C2} & \textbf{C3} & \textbf{C4-L} & \textbf{C4-Q} \\
\midrule

Source
& SWE-agent~\cite{sweagentpaper}
& SWE-rebench~\cite{swerebench}
& LiveClawBench~\cite{liveclawbench}
& \multicolumn{2}{c}{Latent Programming Horizons~\cite{silva2026latentprogramminghorizonscoding}} \\

Scaffold
& SWE-agent
& OpenHands
& OpenClaw
& -- 
& -- \\

Trajectories
& 80{,}036
& 67{,}074
& 6{,}834
& 5{,}000
& 5{,}000 \\

Tasks
& 3{,}591
& 6{,}306
& 134
& 500
& 500 \\

Model(s)
& 3 (Llama-3 8B--405B)
& Qwen3-Coder-480B
& 17
& Laguna-XS.2
& Qwen3.6-35B-A3B \\

Attempts/task
& --
& --
& --
& 10
& 10 \\

Failure rate
& 0.833
& 0.521
& 0.582
& --
& -- \\

Mixed-outcome
& 22.0\%
& 27.6\%
& 21.9\%
& 40.6\%
& 34.8\% \\

Same-unit weight $w$
& $2.5{\times}10^{-4}$
& $3.4{\times}10^{-5}$
& $8.8{\times}10^{-5}$
& --
& -- \\

Mixed units
& 926
& 1{,}741
& 500
& 203
& 174 \\

Within-unit pairs
& 220{,}055
& 38{,}576
& 1{,}000
& --
& -- \\

\bottomrule
\end{tabular}
\end{table*}

C1 contains SWE-agent coding trajectories from three Llama-3 models. C2 was generated with OpenHands v0.54.0 using a single Qwen3-Coder-480B model; its $6{,}306$ tasks span $1{,}823$ repositories. C3 contains trajectories from $17$ models across ten domains. Of its $6{,}834$ recorded trajectories, $6{,}727$ can be parsed into $2{,}269$ units; $488$ of the $500$ mixed units remain parseable and are used for analyses requiring parsed trajectories.
C3 additionally provides difficulty tiers and domain labels assigned by the benchmark authors rather than derived from our outcomes. Failure rates increase from $0.313$ to $0.710$ to $0.962$ across the three tiers, consistent with their intended ordering. We therefore use these labels as outcome-independent task-level information available before execution.


The hidden-state analysis requires access to the model that produced each trajectory, which is unavailable for most public agent corpora. We therefore use the \textsc{Latent Programming Horizons} corpus~\cite{silva2026latentprogramminghorizonscoding} as C4. It contains SWE-bench trajectories from two open mixture-of-experts models: \textsc{Laguna-XS.2} (C4-L) and \textsc{Qwen3.6-35B-A3B} (C4-Q). Both provide ten attempts for each of the same 500 SWE-bench tasks.
\textsc{Laguna-XS.2} has $40$ layers with $d_{\mathrm{model}}=2048$. Each C4 record stores the exact token identifiers consumed by the model and message offsets, allowing us to replay the original token sequence directly rather than reconstructing the chat template.
For hidden-state probing, C4-L retains $142$ mixed units after applying a $16{,}000$-token eligibility criterion, while C4-Q retains all $174$ mixed units. The threshold is imposed for activation-memory feasibility rather than as a modelling choice. Runs above the threshold are excluded whole rather than truncated, preventing the observed prefix from becoming a function of final trajectory length.

\subsection{Evaluation Metrics}
\label{sec:metrics}

We evaluate predictors using pooled and within-task AUROC. Pooled AUROC, \pooled{}, compares successful and failed runs drawn from the full evaluation set. Within-task AUROC, \within{}, instead compares successful and failed runs from the same unit. For unit $u$, let $S_u$ and $F_u$ denote its successful and failed runs and let $A_u$ be its within-unit AUROC. We define

\begin{equation}
\within =
\frac{\sum_u |S_u||F_u| A_u}
     {\sum_u |S_u||F_u|}.
\end{equation}

Only units containing both a success and a failure contribute to \within{}. A predictor that is constant within a unit therefore receives $\within=0.5$.

\paragraph{\textbf{Pooled AUROC decomposition.}}
\label{sec:decomp}
The positive--negative pairs underlying pooled AUROC can be partitioned into same-unit and cross-unit pairs. Let $w$ denote the fraction of all such pairs that come from the same unit. Then

\begin{equation}
\pooled = w\within + (1-w)\cross,
\qquad
w =
\frac{\sum_u P_uN_u}
     {(\sum_u P_u)(\sum_u N_u)},
\end{equation}

where $P_u$ and $N_u$ are the numbers of failed and successful runs in unit $u$, respectively, and \cross{} is AUROC over cross-unit pairs. We use $w$ to quantify how much of pooled AUROC is determined by comparisons within the same model--task unit.

Cross-unit pairs need not be cross-task: they may differ in task, acting model, or both. For single-model corpora (C2, C4-L, and C4-Q), units and tasks coincide. For the multi-model corpora C1 and C3, we report the same-unit weight and separately examine task- and model-level contributions rather than equating cross-unit variation with task variation. For the comparison of task-history and trajectory-based predictors, we additionally report Brier score and log loss to assess probabilistic prediction quality~\cite{Gneiting01032007}; lower values are better.

\subsection{Prediction Conditions}
\label{sec:conditions}

We vary the information available to each predictor, separating information that is constant within a unit from information that changes across repeated attempts.

\paragraph{\textbf{Run-Blind Predictors}}

Our main control is a predictor that never observes the current trajectory. The \emph{difficulty oracle} assigns each run the failure rate of its unit. In multi-model corpora, this oracle reflects the baseline difficulty of the model-task unit and may therefore contain both task- and model-level variation. For pooled evaluation, that rate is computed with the current run left out so that its own outcome cannot predict itself. For within-task evaluation we use the ordinary unit mean, which is constant across runs and therefore yields $\within=0.5$.
We also evaluate deployable run-blind predictors that do not use target-unit outcome history. On C1 and C2, a pre-execution predictor uses issue-text TF-IDF features~\cite{salton1988term} and repository identity, with tasks held out across folds. On C3, we evaluate three run-blind conditions: task metadata (difficulty tier and domain), acting-model identity, and their combination. All are available before execution.

\paragraph{\textbf{Trajectory Predictors}}

Trajectory predictors use the first $k$ turns of the current execution. Behavioral features include turn counts, message and observation lengths and their dispersion, error tokens, repetition, tool diversity, and TF-IDF text features~\cite{salton1988term}. We fit logistic regression~\cite{cox1958regression} and gradient boosting~\cite{friedman2001greedy} and generate out-of-fold predictions using \textsc{GroupKFold}~\cite{pedregosa2011scikit} grouped by task. Thus, a task never appears in both training and test portions of a fold. We compute \pooled{} and \within{} from the same cross-fitted predictions so that the estimands differ only in the pairs being compared. C3 supports an additional decomposition of task history and current-run information. $M_0$ uses only the unit's failure rate estimated from its other runs, $M_1$ uses only the current trajectory, and $M_2$ combines both. The difference between $M_2$ and $M_0$ therefore measures what the current trajectory adds once outcomes from previous attempts on the same model--task unit are known.

\paragraph{\textbf{Published and Released Systems}}

To determine whether our findings are specific to our predictors, we also evaluate systems developed independently of our pipeline. We reimplement Automata~\cite{automata} because it reports the highest pooled AUROC in our survey ($0.941$) and is evaluated on SWE-agent (C1), allowing a same-corpus comparison. We first verify that our reconstruction matches its published SWE-agent result ($0.799$) on complete trajectories, then evaluate fixed prefixes under both \pooled{} and \within{}. The reconstruction uses the published activity-abstracted automaton with gradient boosting; additional implementation checks are reported in the supplementary material~\cite{artifact}.
We also evaluate released systems directly. EarlyEval~\cite{earlyeval} is particularly relevant because it targets early failure prediction, reports $13$--$26\%$ fewer execution steps at $89$--$97\%$ accuracy, and requires no reconstruction. We run its released features on the pre-registered five-fold C2 split, where each of the $6{,}306$ instances appears in test data once. Its step-$0$ checkpoint predicts before the agent acts, providing a direct measure of pre-execution discrimination. We additionally evaluate the released one-class monitor of \citet{realtimerepair} on the same C2 folds and run EarlyEval on C1's three Llama-3 models to test whether the pattern extends across acting models.

\subsection{Hidden-State Probes}
\label{sec:probe}

Observable trajectory features may miss outcome information encoded internally by the acting model. C4 allows us to test this possibility because its weights and exact trajectory tokens are available for replay.
We cut each trajectory after assistant turn $k\in\{1,3,5,10\}$ and extract the hidden state of the last emitted token from layers $\{10,20,30,40\}$, sampling the 40-layer model at approximately quarter-depth intervals. Each retained prefix is replayed in full. We fit an $\ell_2$-regularized linear logistic probe to the raw $2048$-dimensional activation. The probe is deliberately linear so that we test whether outcome information is directly recoverable from the representation without introducing substantial additional predictive capacity.

Each probe uses \textsc{GroupKFold} grouped by task. Regularization strength is selected with a nested \textsc{GroupKFold} inside each training fold and never using test data. We compare these probes with a three-feature length baseline consisting of absolute token position, its logarithm, and assistant-turn count.
We validate the probing pipeline in four ways: (1) we verify model loading with teacher-forced accuracy; (2) replay stored token identifiers directly; (3) compare the within-task AUROC estimator with an independent implementation to $10^{-9}$ agreement; and (4) evaluate positive and negative controls using an outcome-encoding feature, Gaussian noise, and shuffled labels. Numerical validation results are reported with the probe results.

\subsection{Prefix Definition and Leakage Control}
\label{sec:leak}

All trajectory-based predictors satisfy the same deployment constraint: they may use only information available when the prediction is made. In particular, final trajectory length is unknown while an execution is still running.
We therefore define checkpoints by absolute turn count rather than by a fraction of the completed trajectory. Fractional checkpoints depend on final length $T$ and therefore introduce information unavailable online. For Automata, we evaluate both its original fractional checkpoint and deployment-valid checkpoints based on observed turns and tokens, allowing us to quantify the effect of this leakage.
Later checkpoints create a separate selection issue because runs that terminate early disappear from evaluation. To distinguish increasing predictive signal from changing cohort composition, we additionally evaluate a fixed C1 cohort containing only trajectories that reach at least ten turns, so the same runs are compared across early checkpoints.

\subsection{Analysis by Research Question}
\label{sec:analysis}

\paragraph{\textbf{RQ1: Pooled versus Within-Unit Evaluation}}

RQ1 first computes the same-unit pair weight $w$ from Section~\ref{sec:decomp}, which measures the share of positive--negative comparisons underlying pooled AUROC that come from the same unit. We then evaluate the difficulty oracle, deployable pre-execution predictors, and the C3 run-blind conditions under both \pooled{} and \within{}, comparing them with trajectory-based predictors. This determines how much pooled discrimination can be obtained without observing the current execution and how much remains once comparisons are restricted to repeated attempts from the same model--task unit.

\paragraph{\textbf{RQ2: Run-Level Discrimination}}

RQ2 evaluates \within{} at online-computable prefixes for behavioral predictors, published and released systems, and hidden-state probes. On C3, the $M_0$, $M_1$, and $M_2$ conditions quantify the incremental information provided by the current trajectory after accounting for previous outcomes from the same unit.
We additionally examine trajectory depth and acting-model capability. For depth, we use the fixed C1 cohort from Section~\ref{sec:leak} and evaluate a pre-registered same-unit conditional-logit ranker trained directly on within-unit comparisons. For capability, we compare within-task discrimination across C3's $17$ models and within same-family model pairs. Because models differ in factors other than capability, this analysis is interpreted as associative rather than causal.

\paragraph{\textbf{RQ3: Deployment Decisions}}

RQ3 evaluates whether task-level and run-level information support different decisions under a fixed compute budget. We use replay so that both cost and outcome come from recorded executions. At each decision, a policy selects a task and samples uniformly with replacement from its recorded runs. The sampled run contributes its recorded cost and outcome, and solved tasks are removed from further selection. The primary outcome is the number of distinct tasks solved within the budget.
We compare round-robin, static difficulty ranking transferred from another model, and Budgeted Thompson Sampling~\cite{xia2015thompsonsamplingbudgetedmultiarmed}. For each cost-aware policy, the mean token cost of a task is computed once from all recorded runs of the target model and held fixed throughout replay. The static transferred ranking orders tasks by donor-model success rate divided by target model mean cost. Thompson Sampling treats each task as a Beta- Bernoulli arm and selects the task with the largest sampled success probability divided by its target-model mean cost. The uninformative prior is
$\mathrm{Beta}(1,1)$. For transferred priors,

\begin{equation}
\alpha_u = 1 + \lambda s^{\mathrm{other}}_u,
\qquad
\beta_u = 1 + \lambda
\left(n^{\mathrm{other}}_u-s^{\mathrm{other}}_u\right),
\end{equation}

where $s^{\mathrm{other}}_u$ and $n^{\mathrm{other}}_u$ are successful and total runs from the donor model. We use $\lambda=1$ on C4 and $\lambda=0.5$ on C2; neither value is selected on evaluation data, and sensitivity to $\lambda$ is reported in the supplementary material. Allocation results use $200$ budget realizations per setting.

For reference, an oracle policy selects the unsolved task with the highest target-model success rate divided by target-model mean cost, with both quantities computed from the complete target run set. It therefore differs from the non-oracle policies by knowing the target task success rates in advance. Because target mean costs are computed from the complete replay corpus, the cost-aware policies are not directly deployable as cold-start policies. We use the same target-side cost data for all cost-aware policies to isolate differences in task-success information.

We then compare allocation with run-level aborting on C4-L under a held-out protocol. Tasks are split evenly into tuning and evaluation sets, giving $250$ evaluation tasks. Abort thresholds are tuned separately for each policy and budget using only the tuning set. Final evaluation uses $150$ paired replay trials with matched random seeds. The allocation prior is transferred from C4-Q and never uses outcomes from the replayed C4-L tasks. The aborting policy uses the cross-fitted Automata predictor at $k=5$. If a run is stopped, we charge only the tokens consumed through the stopping point, so a run that would later succeed incurs no further cost; this convention favors the aborting policy. We compare round-robin, static transferred ranking, Budgeted Thompson Sampling with uninformative or transferred priors, abort-only, and allocation with monitoring.

Finally, we perform a pre-registered semi-synthetic sweep over monitors with controlled \within{} at $k=5$. Using the same held-out threshold-selection procedure, we determine how much within-task discrimination is required before adding an abort decision improves the number of tasks solved. C4 is the primary replay corpus because it records token cost directly. We repeat the C2 allocation analysis using trajectory length as cost and task-level information transferred from C1.

\subsection{Statistical Analysis}
\label{sec:stats}

Repeated runs from the same $(\text{model},\text{task})$ unit are dependent, so the unit is our primary clustering and resampling level.

\paragraph{\textbf{Uncertainty and between-unit variation.}}
We report $95\%$ bootstrap confidence intervals by resampling units rather than individual runs. For the acting-model capability analysis, we instead resample models. We also report the intraclass correlation (ICC), with a finite-repeat correction for small unit sizes, as a descriptive measure of how much outcome variation lies between units~\cite{icc}.

\paragraph{\textbf{Early-prefix tests.}}
At intervention-relevant prefixes, we test $H_0\!:\within=0.5$ using within-unit permutation tests on out-of-fold predictions. Outcomes are shuffled within each unit while preserving its numbers of successful and failed runs. Non-significant results are interpreted as limited detectable signal rather than evidence that the true effect is exactly zero.

We additionally report the standard deviation of the permutation null,

\begin{equation}
\frac{
\sqrt{\sum_u P_uN_u(P_u+N_u+1)/12}
}{
\sum_u P_uN_u
},
\end{equation}

where $P_u$ and $N_u$ are the failed and successful runs in unit $u$. This quantity summarizes the statistical resolution supported by the observed repeat structure.

\paragraph{\textbf{Resolution and sensitivity analyses.}}
For the C4 probe, we convert uncertainty in \within{} into a minimum detectable effect. At $80\%$ power for a two-sided test with $\alpha=0.05$, the boundary is $0.5+2.80\,\sigma$. On C4-L, $\sigma=0.0193$ gives a resolution boundary of $0.554$, allowing us to distinguish weak early signal from evidence of its complete absence.
We additionally test robustness to alternative weighting and label noise. \within{} is recomputed under unit-uniform macro averaging, task-demeaned scoring, minimum-repeat thresholds, and posterior-predictive weighting over units. Label-noise sensitivity is calibrated by corrupting known fractions of success labels, refitting the predictor, and measuring attenuation. Full sensitivity grids are reported in the replication package.

\paragraph{\textbf{Predictor comparisons and multiple testing.}}
When predictors are evaluated on the same runs, we use shared bootstrap resamples to estimate AUROC differences and confidence intervals. Families of within-task hypothesis tests use Holm--Bonferroni correction at $\alpha=0.05$~\cite{holm}.

\section{Results}
\label{sec:results}

This section presents the results for RQ1 and RQ2. Section~\ref{sec:res-rq1} examines what pooled AUROC measures, 
Section~\ref{sec:res-rq2} evaluates how much run-level signal is available during execution. RQ3, which concerns deployment decisions, is evaluated separately in Section~\ref{sec:evaluation}.

\subsection{RQ1: What Pooled AUROC Measures}
\label{sec:res-rq1}

\begin{table}[t]
\caption{C1 results at fixed-turn checkpoints (Section~\ref{sec:leak}). $n$ is the number of evaluated trajectories, $A_{\mathrm{pool}}$ and $A_{\mathrm{within}}$ are pooled and within-task AUROC, \textit{gap} is their difference, and \textit{units} is the number of mixed model--task pairs.}
\label{tab:c1}
\small
\centering

\begin{tabular}{@{}lrrrrr@{}}
\toprule
Condition & $n$ & \pooled & \within & gap & units \\
\midrule
$k=2$  & 79{,}929 & 0.6435 & 0.5169 & $+0.127$ & 926 \\
$k=5$  & 77{,}951 & 0.6863 & 0.5428 & $+0.144$ & 920 \\
$k=10$ & 63{,}657 & 0.7219 & 0.5987 & $+0.123$ & 885 \\
$k=20$ & 35{,}554 & 0.7463 & 0.6527 & $+0.094$ & 620 \\
\midrule
Difficulty oracle & 80{,}036 & 0.9454 & 0.5000 & $+0.445$ & 926 \\
\bottomrule
\end{tabular}
\end{table}

Pooled AUROC is computed from failed--successful run pairs. We call a pair  \emph{same-unit} when both runs come from the same model--task pair. These same-unit pairs directly test whether a predictor can distinguish successful from failed runs of the same task and model.

These pairs make up only a tiny fraction of pooled AUROC. On C2 and C4-L, the same-unit pair weights are $3.4\times10^{-5}$ and $6.8\times10^{-4}$, respectively. Thus, more than $99.93\%$ of pooled comparisons are between different units. Because C2 and C4-L each contain only one acting model, different units also correspond to different tasks in these corpora. Their pooled AUROC is therefore dominated by cross-task comparisons. Figure~\ref{fig:pooled-within} summarizes representative conditions from RQ1 and RQ2 and relates their within-task discrimination to the deployment threshold estimated in RQ3.

%
\definecolor{oiGreen}{HTML}{009E73}
\colorlet{pwPool}{black!80}
\colorlet{pwWithin}{oiGreen!80!black}
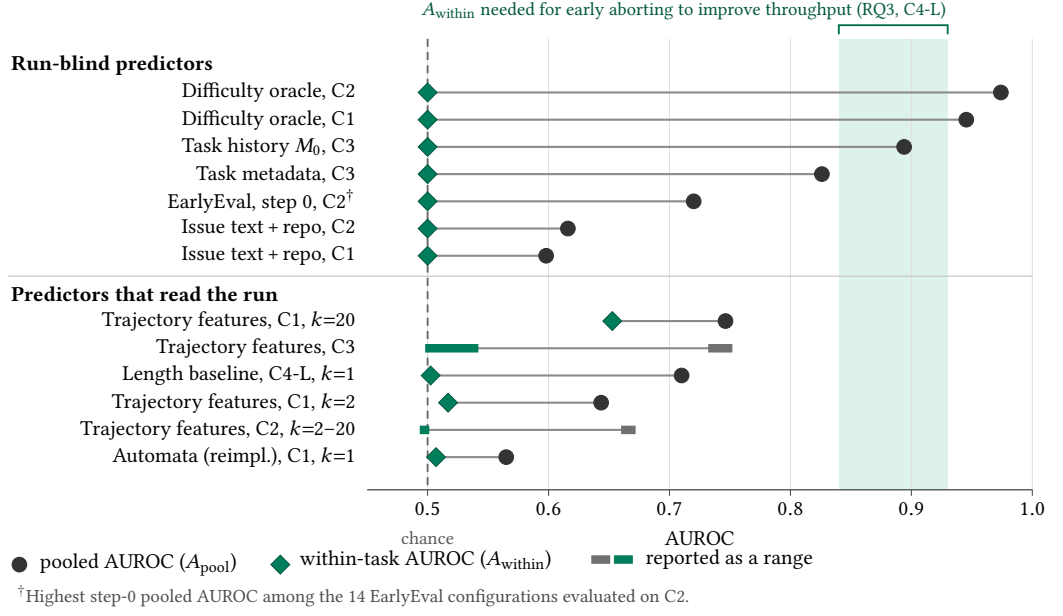
\begin{figure}[t]
\centering
\begin{tikzpicture}[
  font=\footnotesize,
  declare function={ax(\v)=4.75+(\v-0.45)*16.0;},  
  pool/.style={circle, fill=pwPool, draw=pwPool, inner sep=0pt, minimum size=5.5pt},
  wthn/.style={rectangle, rotate=45, fill=pwWithin, draw=pwWithin!70!black, inner sep=0pt, minimum size=5pt},
  conn/.style={black!45, line width=0.8pt},
  rangeP/.style={pwPool!70, line width=3.2pt},
  rangeW/.style={pwWithin, line width=3.2pt},
  lab/.style={anchor=east, inner sep=0pt},
  grp/.style={anchor=west, inner sep=1pt, fill=white, font=\footnotesize\bfseries},
]
\def\dy{0.36}
\newcommand{\pw}[4]{
  \node[lab] at (4.6,-#1*\dy) {#2};
  \draw[conn] ({ax(#4)},-#1*\dy) -- ({ax(#3)},-#1*\dy);
  \node[wthn] at ({ax(#4)},-#1*\dy) {};
  \node[pool] at ({ax(#3)},-#1*\dy) {};}
\newcommand{\pwr}[6]{
  \node[lab] at (4.6,-#1*\dy) {#2};
  \draw[conn] ({ax((#5+#6)/2)},-#1*\dy) -- ({ax((#3+#4)/2)},-#1*\dy);
  \draw[rangeW] ({ax(#5)},-#1*\dy) -- ({ax(#6)},-#1*\dy);
  \draw[rangeP] ({ax(#3)},-#1*\dy) -- ({ax(#4)},-#1*\dy);}

\def\ytop{0.35}
\def\ybot{-15.6*\dy}
\fill[oiGreen!13] ({ax(0.84)},\ytop) rectangle ({ax(0.93)},\ybot);
\foreach \t in {0.5,0.6,0.7,0.8,0.9,1.0}
  \draw[black!12, line width=0.4pt] ({ax(\t)},\ytop) -- ({ax(\t)},\ybot);
\draw[black!60, line width=0.7pt, dash pattern=on 3pt off 2pt] ({ax(0.5)},\ytop) -- ({ax(0.5)},\ybot);
\draw[black!70, line width=0.5pt] ({ax(0.45)},\ybot) -- ({ax(1.0)},\ybot);
\foreach \t in {0.5,0.6,0.7,0.8,0.9,1.0}{
  \draw[black!70] ({ax(\t)},\ybot) -- ++(0,-0.08);
  \node[anchor=north, inner sep=2pt] at ({ax(\t)},\ybot-0.08) {\pgfmathprintnumber[fixed,precision=1,fixed zerofill]{\t}};}
\node[anchor=north] at ({ax(0.725)},\ybot-0.42) {AUROC};
\node[anchor=north, font=\scriptsize, text=black!65] at ({ax(0.5)},\ybot-0.42) {chance};

\draw[oiGreen!70!black, line width=0.7pt] ({ax(0.84)},\ytop+0.06) -- ++(0,0.12) -| ({ax(0.93)},\ytop+0.06);
\node[anchor=south east, inner sep=1pt, font=\scriptsize, text=oiGreen!55!black, align=right]
  at ({ax(0.93)+0.02},\ytop+0.22) {\within{} needed for early aborting to improve throughput (RQ3, C4-L)};

\node[grp] at (0,-0*\dy) {Run-blind predictors};
\pw{1}{Difficulty oracle, C2}{0.974}{0.500}
\pw{2}{Difficulty oracle, C1}{0.9454}{0.500}
\pw{3}{Task history $M_0$, C3}{0.894}{0.500}
\pw{4}{Task metadata, C3}{0.826}{0.500}
\pw{5}{EarlyEval, step 0, C2$^{\dagger}$}{0.72}{0.500}
\pw{6}{Issue text\,+\,repo, C2}{0.616}{0.500}
\pw{7}{Issue text\,+\,repo, C1}{0.598}{0.500}
\node[grp] at (0,-8.4*\dy) {Predictors that read the run};
\pw{9.4}{Trajectory features, C1, $k{=}20$}{0.7463}{0.6527}
\pwr{10.4}{Trajectory features, C3}{0.732}{0.752}{0.498}{0.542}
\pw{11.4}{Length baseline, C4-L, $k{=}1$}{0.710}{0.5025}
\pw{12.4}{Trajectory features, C1, $k{=}2$}{0.6435}{0.5169}
\pwr{13.4}{Trajectory features, C2, $k{=}2$--$20$}{0.660}{0.672}{0.4935}{0.5014}
\pw{14.4}{Automata (reimpl.), C1, $k{=}1$}{0.565}{0.507}
\draw[black!25, line width=0.4pt] (0,-7.75*\dy) -- ({ax(1.0)},-7.75*\dy);

\def\ly{\ybot-0.99}
\node[pool] (m1) at (0.15,\ly) {};
\node[anchor=west] (t1) at (m1.east) {\,pooled AUROC (\pooled{})};
\node[wthn] (m2) at ($(t1.east)+(0.45,0)$) {};
\node[anchor=west] (t2) at ($(m2.east)+(0.02,0)$) {\,within-task AUROC (\within{})};
\draw[rangeP] ($(t2.east)+(0.4,0)$) -- ++(0.25,0) coordinate (r1);
\draw[rangeW] ($(r1)+(0.05,0)$) -- ++(0.25,0) coordinate (r2);
\node[anchor=west] at (r2) {\,reported as a range};
\def\lyy{\ly-0.03}
\node[anchor=north west, font=\scriptsize, text=black!70] at (0,\lyy-0.12)
  {$^{\dagger}$Highest step-0 pooled AUROC among the 14 EarlyEval configurations evaluated on C2.};
\end{tikzpicture}
\caption{\textbf{Pooled versus within-task AUROC across representative prediction conditions.} Run-blind predictors can achieve high pooled discrimination while remaining at chance within task, whereas predictors using current-run information show substantially weaker within-task discrimination. The shaded band marks the approximate within-task AUROC required for early aborting to improve throughput on C4-L.}
\label{fig:pooled-within}
\end{figure}

C3 contains $17$ acting models, so cross-unit comparisons can differ in task, model, or both. We therefore separate these sources. Task metadata alone reaches pooled AUROC $0.826$, while model identity alone reaches $0.568$. Combining the two increases AUROC only to $0.845$, so the cross-unit discrimination on C3 comes from differences between tasks rather than differences between models.

Table~\ref{tab:c1} shows what this means for prediction performance. On C1, the trajectory predictor reaches within-task AUROC $0.5169$ after two turns and $0.5428$ after five turns, while its pooled AUROC is substantially higher. In contrast, the difficulty oracle, which does not observe the current run, reaches pooled AUROC $0.9454$ but exactly $0.5000$ within task. A deployable run-blind predictor based on issue text and repository identity shows the same pattern, reaching pooled AUROC $0.598$ on C1 and $0.616$ on C2 while remaining at $0.500$ within task.

C3 again shows the same pattern. A per-model task oracle reaches a median pooled AUROC of $0.985$, whereas trajectory predictors reach only $0.498$-$0.542$ within task despite pooled AUROC of $0.732$-$0.752$. Approximately $70\%$ of C3's outcome variance lies between units, although this is descriptive rather than a bound on achievable run-level prediction.

\paragraph{\textbf{Does the effect extend beyond coding corpora?}} Because three of our four corpora are software engineering benchmarks, we test whether the same structural pattern holds on tau2-bench~\cite{barres2025tau}, a non-coding tool-use benchmark. We compute two run-blind references. The \emph{difficulty-only ceiling} is an analytic pooled-AUROC ceiling computed from each unit's failure rate using the complete outcome table; it therefore uses the scored run's outcome and is not a deployable predictor. Same-unit comparisons are ties, giving within-task AUROC $0.5$. The \emph{leave-one-out oracle} instead scores each run using the failure rate of the unit's other runs; units with only one run fall back to the global failure rate.

At the domain level, across four domains, $2{,}708$ units, and $10{,}832$ runs, mixed-outcome rates ($28.9\%$--$40.8\%$), same-unit weights ($2.2\times10^{-4}$--$1.6\times10^{-3}$), and difficulty-only ceilings ($0.941$--$0.976$) are comparable to those in our coding corpora. For a finer-grained analysis, we form $26$ cells defined by domain, acting model, and agent configuration and retain the $18$ cells with at least $30$ mixed-outcome units. The retained cells cover retail, telecom, and telecom-workflow; all airline cells fall below this threshold. Across these $18$ cells, difficulty-only ceilings range from $0.927$ to $0.975$ (median $0.954$), while leave one-out oracle AUROCs range from $0.752$ to $0.901$ (median $0.855$).

\rqbox{Answer to RQ1.}{
Pooled AUROC is dominated by comparisons between different model--task units. In our corpora, same-unit comparisons contribute no more than $6.8\times10^{-4}$ of the pair mass. As a result, a predictor can achieve high pooled AUROC by identifying difficult tasks without distinguishing successful from failed runs of the same task. This is visible in the difficulty oracle, which reaches pooled AUROC $0.9454$ on C1 while remaining at chance ($0.5000$) within task. Pooled AUROC therefore does not by itself establish run-level failure discrimination.
}

\subsection{RQ2: Run-Level Information at Deployable Prefixes}
\label{sec:res-rq2}

\paragraph{\textbf{Trajectory-Based Predictors}}

We first measure the incremental value of the current trajectory using the predictors from Section~\ref{sec:conditions}. On C3, with $6{,}727$ runs and $2{,}269$ units, $M_0$ achieves pooled AUROC, Brier score, and log loss of $0.894$, $0.103$, and $0.350$; the corresponding values are $0.726$, $0.211$, and $0.606$ for $M_1$, and $0.923$, $0.103$, and $0.340$ for $M_2$. Thus, the task-level predictor alone already falls within the published AUROC range of $0.85$--$0.94$.

Because $M_0$ uses previous attempts from the same unit, adding the full current trajectory increases pooled AUROC by only $0.029$, improves log loss by $0.0101$, and changes Brier score by only $+0.0007$. Most of the gain therefore comes from ranking rather than probability estimation. Within task, $M_0$ is $0.5$ by construction, while trajectory features alone reach $0.587$; a within-task score for $M_2$ is not identifiable under leave-one-out task scoring.

C2 provides the clearest null result. Across $67{,}074$ trajectories from $1{,}823$ repositories, within-task AUROC remains between $0.4935$ and $0.5014$ from two to twenty turns, with all intervals covering $0.50$. Median unit AUROC is exactly $0.500$, and only $46\%$--$48\%$ of units exceed chance. In contrast, pooled AUROC remains $0.660$--$0.672$, yielding a pooled--within gap of $+0.16$ to $+0.18$, while the difficulty oracle reaches $0.974$.

C1 behaves differently: within-task AUROC increases from $0.517$ to $0.653$, whereas C2 remains near chance ($0.494$--$0.497$). Re-running C1 with C2's eight features changes the curve by at most $0.035$, suggesting that feature choice does not explain the difference. After Holm correction, no permutation test rejects the null that within-task AUROC equals $0.5$; C3 has one nominal result at five turns ($p=0.027$), but it does not survive correction. Nevertheless, C3's task-metadata predictor exceeds the trajectory predictor by $+0.134$ pooled AUROC ($[+0.120,+0.148]$), showing that pooled evaluation can favor a predictor with no run-level discrimination.

\paragraph{\textbf{Hidden-State and Length Signals}}
\label{sec:res-equiv}

We test whether hidden states contain run-specific failure information using cross-fitted probes on C4. The signal remains weak: mean within-task AUROC is $0.509$ on C4-Q, with only one configuration reaching $0.551$ after Holm correction, and $0.513$ on C4-L, where no configuration is significant.

At the same time, hidden states strongly encode trajectory length ($R^2$ up to $0.98$) and repository identity (AUROC up to $1.00$). Control experiments confirm that the probing pipeline can recover genuine signal: an outcome-encoding feature reaches AUROC $1.000$, while noise and shuffled labels remain near chance. Thus, the weak within-task results are not explained by an incapable probe. This contrasts with recent evidence that hidden-state probes can predict eventual failure from the first interaction round~\cite{doomed}. Our findings are consistent with Bulut~\cite{bulut2026s}, whose main reconstruction shows that strong pooled hidden-state prediction drops to near-chance performance after conditioning on problem identity. Together, these results suggest that early hidden-state predictability can reflect task or problem difficulty rather than run-specific outcome information.

\definecolor{oiBlue}{HTML}{0072B2}
\definecolor{oiGreen}{HTML}{009E73}

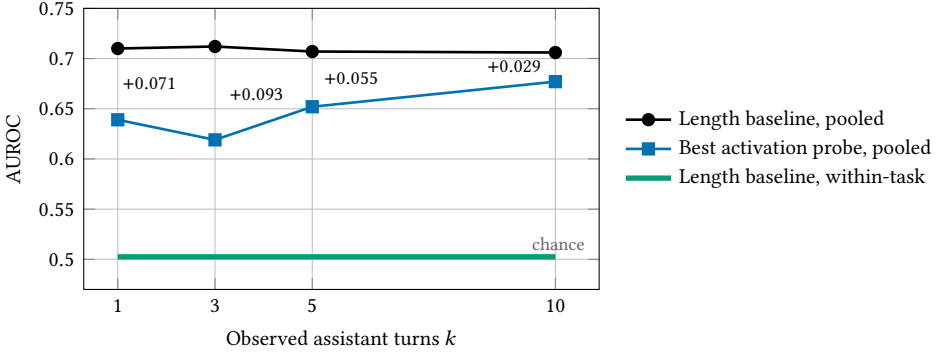
\begin{figure}[t]
\centering

\begin{tikzpicture}
\begin{axis}[
  width=8.4cm,
  height=5.3cm,
  font=\footnotesize,
  xmin=0.3,
  xmax=10.9,
  xtick={1,3,5,10},
  xticklabels={1,3,5,10},
  ymin=0.47,
  ymax=0.75,
  ytick={0.5,0.55,0.6,0.65,0.7,0.75},
  xlabel={Observed assistant turns $k$},
  ylabel={AUROC},
  grid=major,
  legend style={at={(1.03,0.5)},anchor=west,draw=none,font=\footnotesize},
  legend cell align=left,
  clip=false
]

\node[anchor=south east,font=\scriptsize,text=black!60]
  at (axis cs:10.8,0.5) {chance};

\addplot[
  black,
  mark=*,
  line width=1pt
]
coordinates {
  (1,0.710)
  (3,0.712)
  (5,0.707)
  (10,0.706)
};
\addlegendentry{Length baseline, pooled}

\addplot[
  oiBlue,
  mark=square*,
  line width=1pt
]
coordinates {
  (1,0.639)
  (3,0.619)
  (5,0.652)
  (10,0.677)
};
\addlegendentry{Best activation probe, pooled}

\addplot[
  oiGreen,
  line width=2pt
]
coordinates {
  (1,0.5025)
  (10,0.5025)
};
\addlegendentry{Length baseline, within-task}

\node[font=\scriptsize] at (axis cs:1.65,0.675) {$+0.071$};
\node[font=\scriptsize] at (axis cs:3.85,0.665) {$+0.093$};
\node[font=\scriptsize] at (axis cs:5.80,0.680) {$+0.055$};
\node[font=\scriptsize] at (axis cs:9.15,0.692) {$+0.029$};

\end{axis}
\end{tikzpicture}

\caption{Length baseline versus activation probe on C4-L.}
\label{fig:length-hidden-matrix}

\end{figure}

Figure~\ref{fig:length-hidden-matrix} makes this distinction especially clear. Across one, three, five, and ten turns, the three-feature length baseline exceeds the best $2048$-dimensional activation probe in pooled AUROC by $+0.071$, $+0.093$, $+0.055$, and $+0.029$, respectively, yet achieves only $0.502$--$0.503$ within-task AUROC. Extending the observation of \citet{beyondres}, trajectory length therefore remains strongly predictive in pooled evaluation while providing essentially no within-task discrimination once both the acting model and task are fixed. A non-significant result does not imply zero signal, so we quantify the resolution of C4-L. Across $142$ mixed units, the bootstrap standard error is $0.0193$. At $80\%$ power, effects of about $0.054$ are detectable, allowing us to rule out within-task AUROC values of roughly $0.554$ or higher while leaving smaller effects such as $0.53$ possible. The pooled--within gap is $+0.107$ with a $90\%$ interval of $[+0.046,+0.165]$. The pattern is also stable under alternative weighting schemes and plausible label noise: C1 remains above chance, C2 remains near chance, and the observed mislabelling rate is insufficient to hide effects near $0.60$ or explain the pooled--within gap. Thus, modest run-level signal cannot be excluded, but strong early discrimination is inconsistent with the observed data.

\paragraph{\textbf{Published and Released Monitors}}
\label{sec:res-automata}
\label{sec:res-released}

To test whether stronger existing methods recover run-level signal that our predictors miss, we first reimplement Automata. On SWE-agent, the original work reports pooled AUROC $0.799$, while our reconstruction reaches $0.806$. Despite reproducing the published result, early within-task discrimination remains weak: at one turn, Automata reaches $0.507$ within-task AUROC versus $0.565$ pooled.

We also find that fractional checkpoints introduce look-ahead leakage because they depend on final trajectory length. At the $25\%$ checkpoint, Automata reaches $0.6297$, compared with $0.5658$ at a fixed-turn checkpoint observing a similar number of turns. Final trajectory length alone reaches $0.6138$, while fixed turn and token checkpoints yield similar scores ($0.5658$ and $0.5710$), indicating that much of the fractional-checkpoint gain comes from leaked length information. To rule out implementation effects, we next evaluate released systems directly. On C2, EarlyEval reaches pooled AUROC up to $0.72$ before the agent acts, while within-task AUROC is exactly $0.500$ across all model--fold combinations. After execution begins, within-task AUROC across $14$ variants remains $0.49$--$0.52$ after one step and $0.45$--$0.52$ through five steps. Even at run end, within-task AUROC stays below 0.58 while pooled AUROC reaches $0.77$, with 99.9\% of comparisons being cross-unit.

The released one-class monitor of \citet{realtimerepair} likewise shows weak transfer on C2, with pooled AUROC of $0.44$--$0.55$ through thirty steps. Its later signal is modest, reaching pooled AUROC $0.58$--$0.60$ and within-task AUROC $0.55$--$0.57$. EarlyEval on C1 again shows stronger run-level signal than on C2. Pre-run within-task AUROC is $0.500$, rising to $0.49$--$0.56$ after one step and $0.50$--$0.57$ after five. Most mixed test units, however, come from one model, and later checkpoints contain too few units for reliable interpretation.

\paragraph{\textbf{When Does Run-Level Signal Emerge?}}
\label{sec:res-mechanism}

Two factors explain much of the variation across corpora: trajectory depth and acting-model capability.

First, run-level signal increases with depth. On a fixed C1 cohort of $63{,}657$ traces that all reach at least ten turns, within-task AUROC rises from $0.509$ at one turn to $0.598$ at ten, showing that the increase is not caused by survivor selection. Training directly for same-unit ranking improves early discrimination only modestly, reaching $0.526$, $0.535$, and $0.547$ at one, three, and five turns, still far below the intervention requirements in Section~\ref{sec:eval-requirement}. This depth dependence helps reconcile our findings with FailFast--RestartSmart~\cite{wang2026failfastrestartsmartearlyfailure}, which reports effective failure-triggered termination on SWE-bench Verified but suppresses alarms during the first $20$ steps and augments terminal supervision with intermediate test-progress signals. A similar but weaker depth-dependent pattern appears in Bulut~\cite{bulut2026s}: their main reconstructed hidden-state probe remains near chance within problem across checkpoints, while a post-hoc problem-centered probe recovers modest within-problem signal at later positions. Together, these results suggest that actionable run-specific information may become more available as trajectories develop, although its strength depends on the task, model, and representation.

Second, failures become less predictable as the acting model strengthens. Across C3's $17$ models, within-task AUROC correlates negatively with model capability ($\rho=-0.662$, $p=0.0038$). The weakest model reaches $0.832$, whereas the two strongest reach only $0.444$ and $0.490$. The same pattern appears within model families and on C1, where ten-turn AUROC decreases from $0.636$ for Llama-8B to $0.569$ for Llama-405B. We interpret this as an association rather than a causal effect.

These factors also help reconcile the corpora. C2's stronger model remains near chance across depths ($0.500$--$0.504$), whereas C1's weaker models become more predictable as trajectories develop. C4 shows the same pattern: behavioral features remain near chance at the early prefixes used by the hidden-state probes and reach $0.656$ only on full trajectories. Together, the results suggest that early failures are most legible for weaker agents and become easier to detect later in execution.

\rqbox{Answer to RQ2.}{
Run-specific failure information is weak at early, online-computable prefixes across the predictors and representations we evaluate. Early within-task AUROC is $0.507$ for Automata after one turn, $0.49$--$0.52$ for EarlyEval after one step, and $0.509$ on average for C4-Q hidden-state probes. The signal becomes stronger later in execution and is more pronounced for weaker models: on C1, within-task AUROC rises to $0.598$ at ten turns, while across C3's $17$ models it decreases with model capability ($\rho=-0.662$). Overall, the available evidence supports weak early run-specific discrimination rather than an absence of run-level information altogether; stronger signal becomes detectable as trajectories develop.
}




\section{Evaluation: Deployment Decisions (RQ3)}
\label{sec:evaluation}
\label{sec:res-rq3}

RQ1 and RQ2 examine what failure predictors measure. This section evaluates whether that distinction affects deployment decisions. Using replay under fixed token budgets, we compare task-level allocation with run-level aborting, using recorded costs and outcomes from real executions.

\definecolor{oiBlue}{HTML}{0072B2}
\definecolor{oiSky}{HTML}{56B4E9}
\definecolor{oiGreen}{HTML}{009E73}
\definecolor{oiVerm}{HTML}{D55E00}
\definecolor{oiPurple}{HTML}{CC79A7}
\begin{figure}[t]
\centering
\begin{tikzpicture}
\begin{axis}[
  width=8.6cm, height=6.0cm,
  font=\footnotesize,
  xmin=0.3, xmax=10.7,                
  xtick={1,2.5,5,10}, xticklabels={1M,2.5M,5M,10M},
  ymin=0, ymax=180, ytick={0,40,80,120,160},
  xlabel={Token budget}, ylabel={Distinct tasks solved},
  grid=major, major grid style={black!10},
  axis line style={black!70}, tick style={black!70},
  legend style={at={(1.03,0.5)}, anchor=west, draw=none, font=\footnotesize,
                row sep=1pt, cells={anchor=west}},
  legend cell align=left,
  error bars/error bar style={solid, line width=0.6pt},
  error bars/error mark options={rotate=90, mark size=1.6pt, line width=0.6pt},
  every axis plot/.append style={line width=1.0pt, mark size=2.2pt},
]
\addplot+[oiBlue, solid, mark=square*, mark options={solid, fill=oiBlue},
  error bars/.cd, y dir=both, y explicit] coordinates {
  (0.75,55.2) += (0,2.8) -= (0,3.2)
  (2.25,99.5) += (0,4.5) -= (0,5.5)
  (4.75,134.2) += (0,5.8) -= (0,7.2)
  (9.75,151.7) += (0,7.3) -= (0,10.7)};
\addlegendentry{Static transferred ranking}
\addplot+[oiGreen, solid, mark=*, mark options={solid, fill=oiGreen},
  error bars/.cd, y dir=both, y explicit] coordinates {
  (0.85,52.5) += (0,3.5) -= (0,4.5)
  (2.35,94.4) += (0,5.6) -= (0,6.4)
  (4.85,130.8) += (0,6.2) -= (0,6.8)
  (9.85,161.2) += (0,5.8) -= (0,6.2)};
\addlegendentry{TS\,+\,transferred}
\addplot+[oiPurple!85!black, dash pattern=on 3pt off 2pt, mark=diamond*, mark options={solid, fill=white},
  error bars/.cd, y dir=both, y explicit] coordinates {
  (0.95,48.3) += (0,3.7) -= (0,4.3)
  (2.45,93.8) += (0,5.2) -= (0,5.8)
  (4.95,128.6) += (0,7.4) -= (0,5.6)
  (9.95,159.2) += (0,5.8) -= (0,6.2)};
\addlegendentry{TS\,+\,transferred\,+\,tuned monitor}
\addplot+[oiSky, dash pattern=on 4pt off 1.5pt on 1pt off 1.5pt, mark=triangle*, mark options={solid, fill=oiSky},
  error bars/.cd, y dir=both, y explicit] coordinates {
  (1.05,47.3) += (0,4.7) -= (0,3.3)
  (2.55,80.9) += (0,5.1) -= (0,4.9)
  (5.05,115.3) += (0,5.7) -= (0,5.3)
  (10.05,143.7) += (0,6.3) -= (0,5.7)};
\addlegendentry{TS, uninformative}
\addplot+[oiVerm, dash pattern=on 5pt off 2pt, mark=x, mark options={solid, line width=1.1pt}, mark size=2.8pt,
  error bars/.cd, y dir=both, y explicit] coordinates {
  (1.15,16.8) += (0,7.2) -= (0,6.8)
  (2.65,45.3) += (0,9.7) -= (0,9.3)
  (5.15,82.9) += (0,10.1) -= (0,10.9)
  (10.15,144.2) += (0,9.8) -= (0,9.2)};
\addlegendentry{Tuned abort only}
\addplot+[black!55, dash pattern=on 1pt off 1.5pt, mark=o, mark options={solid, fill=white},
  error bars/.cd, y dir=both, y explicit] coordinates {
  (1.25,16.0) += (0,8.0) -= (0,6.0)
  (2.75,38.6) += (0,9.4) -= (0,10.6)
  (5.25,76.8) += (0,13.2) -= (0,11.8)
  (10.25,146.9) += (0,9.1) -= (0,7.9)};
\addlegendentry{Round-robin}
\end{axis}
\end{tikzpicture}
\caption{Tasks solved on the 250-task C4-L split under fixed token budgets. Points average 150 paired replay trials; bars show 95\% percentile intervals. Abort thresholds are tuned separately. TS denotes budgeted Thompson sampling with task information transferred from C4-Q.}
\label{fig:rq3-budget}
\end{figure}
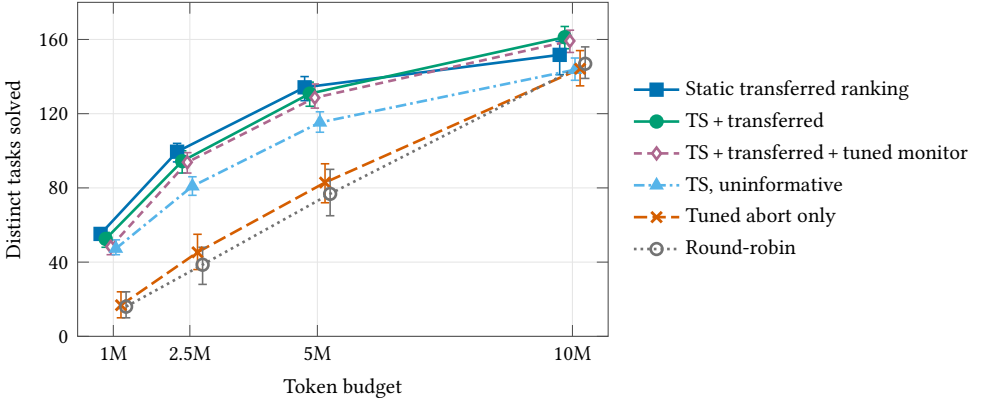

\vspace{-5pt}
\subsection{\textbf{Task-Level Allocation}}
\label{sec:eval-gap}

Task-level allocation is most valuable under tight budgets. On C4-Q, allocation solves $162.0$ tasks at $2$M tokens, compared with $53.0$ for round-robin and $172.9$ for the oracle, reaching $93.7\%$ of the oracle's throughput ($162.0/172.9$).

Prior quality matters. On C2, an uninformative prior outperforms a transferred one ($52.5$ vs.\ $46.1$ tasks at $2$M), consistent with differences between the donor and target settings. Static transferred rankings are competitive at scarce budgets, outperforming adaptive allocation on C4-Q ($166.6$ vs.\ $162.0$ at $2$M) and C4-L ($55.2$ vs.\ $51.9$ at $1$M).

Across the $500$ shared C4 tasks, the per-task success rates of C4-Q and C4-L are strongly correlated (Pearson $r=0.863$). The static transferred score is also strongly correlated with the oracle score (Spearman $\rho=0.916$), although both scores use the same target mean-cost denominator and therefore share cost information.

\subsection{\textbf{Abort or Reallocate}}
\label{sec:eval-abort}

We compare allocation and aborting on C4-L under a held-out protocol. Tasks are split evenly into tuning and evaluation sets, abort thresholds are selected separately for each budget using only the tuning set, and policies use paired replay seeds. The allocation prior is transferred from C4-Q, while the run-level monitor is the five-turn cross-fitted automaton from Section~\ref{sec:res-automata}. Aborted runs are charged only the tokens consumed through the stopping point, which favors the aborting policy. Figure~\ref{fig:rq3-budget} compares allocation and aborting across token budgets on held-out C4-L.

Adding the monitor to budgeted Thompson sampling does not improve throughput: across budgets, the paired difference ranges from $-4.2$ to $-0.6$ tasks, with all $95\%$ intervals spanning zero. Abort-only intervention performs substantially worse than allocation, losing $17.2$--$49.0$ tasks, and does not outperform round-robin despite threshold tuning. As an additional compute-control baseline, turn caps produce the same qualitative result. On C2, capped round-robin solves 15.7 tasks versus 37.9 for allocation at 2M, and adding caps to allocation lowers throughput.

\subsection{\textbf{How Good Would a Monitor Have to Be}}
\label{sec:eval-requirement}

We next ask how accurate a run-level monitor must be before early stopping becomes useful. Using semi-synthetic monitors with controlled within-task AUROC at five turns, we identify when allocation plus monitoring first outperforms allocation alone.

On C4-L, improvement requires about $0.84$ within-task AUROC at a $10$M-token budget and about $0.93$ at $5$M. This range is shown as the shaded band in Figure~\ref{fig:pooled-within}, highlighting the gap between the run-level discrimination achieved by the evaluated monitors and the level required for early aborting to improve throughput. When monitor errors are concentrated on harder tasks, the requirement rises to roughly $0.90$ -$1.00$. On C2, even perfect discrimination does not improve throughput between $2$M and $10$M tokens and becomes beneficial only at $25$M. Early monitors reach only 0.50–0.55 within-task AUROC, well below the level needed for beneficial aborting.

\rqbox{Answer to RQ3.}{Task-level information is substantially more useful for compute allocation than the measured run-level signal is for early aborting under the studied budgets. Static difficulty ranking solves up to $3.5\times$ more tasks than round-robin, whereas adding the measured monitor changes throughput by only $-4.2$ to $-0.6$ tasks. Early aborting becomes useful only around $0.84$--$0.93$ within-task AUROC, far above the observed range of $0.50$--$0.55$.}

\section{Discussion}
\label{sec:discussion}

\paragraph{\textbf{Implications of the findings.}}
Our results separate task-level predictability from run-level predictability. Pooled AUROC is dominated by cross-unit comparisons, allowing predictors that never observe the current run to achieve strong aggregate performance; for example, a task-difficulty scorer reaches $0.9454$ pooled AUROC while remaining at $0.5000$ within task. In contrast, early within-unit discrimination is consistently weak across the monitors and probes we evaluate, and the signal that does emerge tends to appear later and for weaker acting models. This distinction matters operationally: task-level information supports compute allocation, whereas run-level information is required for deciding whether to stop an ongoing execution. Under fixed token budgets, allocation outperforms abort-only strategies, while the monitor does not improve throughput.

\paragraph{\textbf{Implications for evaluation.}}
Pooled AUROC alone is therefore insufficient for evaluating runtime monitors. We recommend reporting within-unit AUROC, the same-unit pair weight, and the pooled--within gap alongside the aggregate score. Evaluation checkpoints should also use only information available during execution; fractional checkpoints based on final trajectory length can leak future information and inflate performance. Finally, intervention policies should be compared against task-level allocation baselines under the same compute budget.

\paragraph{\textbf{Scope of the conclusions.}}
We do not claim that trajectories contain no run-level information or that monitoring is inherently ineffective. Rather, the early signal we observe is weak, and the evaluated monitors remain far below the roughly $0.84$--$0.93$ within-task AUROC at which aborting becomes beneficial in our replay. These thresholds depend on the benchmark and budget, and our deployment results are based on recorded-run replay rather than live intervention. For previously unseen tasks with no execution history, task difficulty must instead be estimated from pre execution task features, transferred from related models or settings, or learned from initial attempts. Our strongest allocation results therefore apply most directly when repeated-attempt information or transferable task difficulty estimates are available.

\section{Threats to validity}
\label{sec:threats}

\paragraph{\textbf{Construct Validity.}}
Our analyses rely on benchmark-provided outcome labels, so label errors could weaken the measured relationship between trajectory information and outcome and pull within-task AUROC toward $0.5$. We therefore test sensitivity to plausible label noise. The observed mislabelling rate is too small to explain the pooled--within gap or to hide effects around $0.6$ or higher, although smaller effects remain possible.

\paragraph{\textbf{Internal Validity.}}
Automata is reconstructed from its published description because the original artifacts are unavailable. We mitigate implementation differences by reproducing its published SWE-agent pooled result and by separately evaluating systems with released implementations. We also use fixed observed-turn checkpoints to avoid leakage from final trajectory length, and a fixed C1 cohort to reduce survivor effects at later checkpoints. Because models in the capability analysis differ in more than capability alone, we interpret that result as associative rather than causal.

\paragraph{\textbf{External Validity.}}
Our main corpora are coding-heavy, although C3 includes non-coding domains. We therefore examine the same structural quantities on four non-coding tau2-bench domains and observe comparable patterns. This supports the broader relevance of the pooled-versus-within issue, but not the generalization of all trajectory-level findings. Hidden-state results are limited to the two open C4 models, and the deployment analysis uses replay rather than live intervention. The allocation results also apply most directly to settings with repeated attempts.

\paragraph{\textbf{Statistical Conclusion Validity.}}
Within-task AUROC is defined only for mixed units, so our estimates are conditional on units containing both successes and failures. Alternative weighting schemes do not materially change the early estimates. Near-chance AUROC should also be interpreted cautiously: failure to reject $H_0:\within=0.5$ does not imply that no run-level signal exists. We therefore describe the early signal as weak rather than absent and report confidence intervals, clustered resampling, within-unit permutation tests, and Holm--Bonferroni correction.
\vspace{-2pt}

\section{Conclusion}
In this paper, we examined what failure-prediction performance measures for LLM agents and whether those predictions support useful deployment decisions. Using benchmarks with repeated attempts of the same model-task unit, we separated task-level discrimination from discrimination between successful and failed runs of the same task. We evaluated run-blind baselines, trajectory predictors, hidden-state probes, and published and released failure-prediction systems, and then used replay experiments to study the consequences for compute allocation.


Our results show that strong pooled AUROC can largely reflect differences between units, and in particular differences in task difficulty, rather than information about the current run.
More than $99.93\%$ of the positive--negative pairs underlying pooled AUROC compare runs from different units, allowing predictors that never observe the current trajectory to achieve strong pooled performance while remaining near chance within task. Early within-task discrimination is generally weak across the evaluated predictors and monitoring systems, although the signal becomes stronger later in execution and is more visible for weaker acting models.

This distinction also matters operationally. Under a fixed token budget, task-level allocation outperforms aborting runs with the evaluated monitors, and adding the monitor to an allocation policy does not improve throughput under the studied conditions. In the C4-L sensitivity analysis, within-task AUROC of approximately $0.84$--$0.93$ is required before aborting becomes beneficial.

Overall, our study shows that failure-prediction evaluation should distinguish task-level predictability from run-level predictability. We recommend reporting within-task AUROC and the same-unit pair weight alongside pooled AUROC, evaluating predictors only at prefixes available during execution, and comparing run-level intervention policies against task-level allocation baselines. Doing so makes clear what information a failure predictor captures and whether that information supports the deployment decision for which the predictor is intended.

\section*{Data Availability}
\label{Repo}
All code, scripts, configurations, and additional artifacts used in this study are publicly available at~\cite{artifact}.

\bibliographystyle{ACM-Reference-Format}
\bibliography{refs}

\end{document}